\pdfoutput=1
\documentclass[11pt]{article}

\usepackage{ACL2023}

\usepackage{times}
\usepackage{latexsym}
\usepackage{graphicx}
\usepackage{tipa}
\usepackage{booktabs}
\usepackage{listings}
\usepackage{multicol}

\usepackage[T1]{fontenc}
\usepackage[utf8]{inputenc}
\newcommand{\zh}[1]{\begin{CJK}{UTF8}{bsmi}#1\end{CJK}}

\usepackage{microtype}

\usepackage{inconsolata}

\usepackage{CJKutf8}

\title{Multi-Dialectal Representation Learning of Sinitic Phonology}

\author{Zhibai Jia \\
  No.2 High School of East China Normal University \\
  \texttt{jiazhibai@proton.me} }

\begin{document}
\maketitle
\begin{abstract}
 Machine learning techniques have shown their competence for representing and reasoning in symbolic systems such as language and phonology. In Sinitic Historical Phonology, notable tasks that could benefit from machine learning include the comparison of dialects and reconstruction of proto-languages systems. Motivated by this, this paper provides an approach for obtaining multi-dialectal representations of Sinitic syllables, by constructing a knowledge graph from structured phonological data, then applying the BoxE technique from knowledge base learning. We applied unsupervised clustering techniques to the obtained representations to observe that the representations capture phonemic contrast from the input dialects. Furthermore, we trained classifiers to perform inference of unobserved Middle Chinese labels, showing the representations' potential for indicating archaic, proto-language features. The representations can be used for performing completion of fragmented Sinitic phonological knowledge bases, estimating divergences between different characters, or aiding the exploration and reconstruction of archaic features.
\end{abstract}

\section{Introduction}

The evolution of languages in the Sinitic family created intricate correspondences and divergences in its dense dialect clusters. Investigating the dynamics of this evolution, through comparison and proto-language reconstruction, is an essential task to Sinitic Historical phonology. However, it may be costly for researchers to manually probe through the groups in search of phonological hints. Hence, it is desirable to accelerate the process with modern algorithms, specifically, representation learning.  

%Several lines of research may benefit from such a tool. In dialectology, there is need for estimating divergence between phonological systems, which can be achieved quantitatively with the representations. In historical phonology, the reconstruction of a proto-language demands deep scrutiny of dialect systems whose efficiency can be improved with manipulating the representations.

Graph-based machine learning \cite{Makarov2021SurveyOG} have gained increasing attention in recent years, due to their versatility with data with flexible structures. Especially, missing link prediction algorithms for knowledge graphs \cite{Wang2021ASO} \cite{Zhu2022MultiModalKG} can uncover a latent structure in noisy and incomplete knowledge. In the case for learning phonological representations, using graph-based learning can allow for more comprehensive integration of multi-dialectal evidence. Thus, we propose applying graph-based techniques for multi-dialectal representation learning.

We construct a knowledge graph from the multi-dialectal phonological data, by abstracting unique phonetic components and individual characters into two kinds of nodes. Then, we connect them with edges specific to the dialect type wherein the character is associated with the given component. On the constructed knowledge graph, we train the BoxE algorithm \cite{abboud_boxe_2020}, a Box Embedding Model for knowledge base completion. Finally, we evaluate the obtained representations with unsupervised and supervised clustering, as well as MLP probes based on Middle-Chinese-derived labels, to show this tool’s value for Sinitic phonological investigation.

\section{Background on Sinitic Languages}

The analysis of Sinitic languages face a few specific challenges due to unique phonological characteristics. These characteristics define crucial details of our design.

In Sinitic languages, morphemes are primarily monosyllabic. Hence, Chinese writing binds one syllable to each of its glyphs, known as characters. A syllable in Sinitic can be decomposed into an initial, a final and a tone. \cite{Shen2020APH} Initials refer to the consonant-like sounds at the beginning of a syllable, which include both stops (e.g. /p-/, /b-/) and fricatives (e.g. /s-/, /\textipa{S}-/). These initials could be combined with various finals to form syllables. Finals refer to the vowel-like sounds at the end of a syllable, which included both simple vowels (e.g. /-a/, /-i/, /-u/), complex vowels (e.g. /-ai/, /-ao/, /-ei/), and vowels combined with consonant codas (/-m/,/-n/,/-\textipa{N}/,/-p/,/-t/,/-k/). Tones refer to the pitch patterns associated with syllables in Chinese. Tones could distinguish between words that were otherwise homophonous, and they were an important part of the Chinese phonological system. 

Due to the early conception of the Chinese writing system, syllables from different Sinitic languages can usually be aligned to each other through a written form. As this alignment is typically implemented in databases of raw Sinitic data, the difficulty of cognate identification is drastically reduced, facilitating analysis. However, the simple syllable structure introduces large amounts of homophones, words sharing same pronunciations, into Sinitic languages. This hinders the use of the comparative method in reconstructing a Sinitic proto-language. The existence of a supersegmental tone feature also complicates a historical analysis of Sinitic languages. 

\begin{figure}[ht]
\centering
\includegraphics[width=0.5\textwidth]{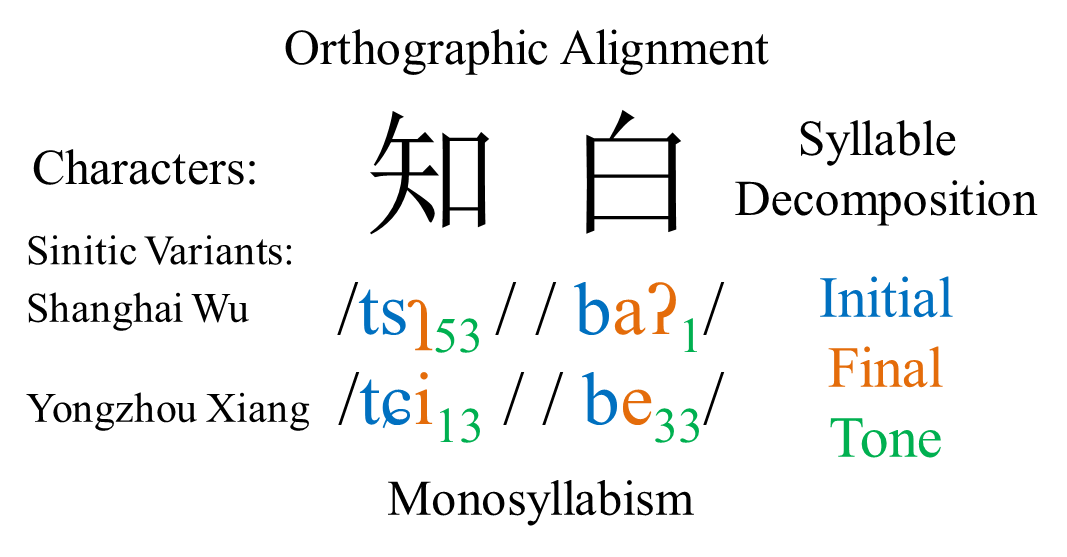}
\caption{Highlighting key characteristics of Sinitic relevant to our approach. Characters are the central identity in the multi-dialectal representations. The orthographic alignment of sub-syllable components form the structure of data used in this study.}
\label{fig:Sinitic}
\end{figure}

Two factors that motivate the use of a graph-based method include the uniform structure of Sinitic syllables and their intimate relationship with characters. The intuitive syllable decomposition and the glyph-based alignment inspire viewing the components contextualized in various dialects as different "observations" of a single character. Theoretically, these observations are derivable from the reading of the character in the proto-language.

\section{Related Work}
The practice of computationally-aided proto-language construction, often associated with cognate identification, has been extensively considered in the past two decades \cite{nerbonne_computing_2007}. Examples include \citep{steiner_pipeline_2011}  which draws insights from bio-informatics and the classical comparative workflow, and \citep{list_potential_2017}, which compared many methods for cognate identification. An relevant insight from the latter paper is that language-specific methods often outperform language-general ones, especially for languages like Sinitic. An epitome of neural methods for proto-language reconstruction would be \cite{meloni-etal-2021-ab}, in which Latin is reconstructed from Romance descendent languages with a encoder-decoder structure. Though, our approach differs from their study in many crucial aspects. In \citealt{meloni-etal-2021-ab}, the reconstruction is supervised, with the proto-language Latin provided at training time. But our method targets not only documented proto-languages like Middle Chinese, but also unknown, intermediate varieties in the development from ancient Sinitic to modern dialects, which requires an unsupervised approach. Additionally, in term of techniques, their use of GRU and attention-based transducers contrasts with our emphasis on a graph-based method.

Considering the representation learning of Sinitic, we found abundant literature on the topic of speech recognition \citep{ma2022survey}, segmentation and synthesis, which often yield representations of certain phonological relevance as by-product. Though, these studies devote heavily to a few major languages, specifically Mandarin or Cantonese, and, since they are rarely claim motivation from historical phonology, seldom take a multi-lingual or multi-dialectal approach.

While speech representation learning often serve the aformentioned purposes, the proposals of using neural networks to model phonetics and phonology from either symbolic abstractions or acoustic data in order to examine theories in these fields are relevant to this study. Unsupervised binary stochastic autoencoders were explored in \citep{shain_measuring_2019}. GAN (Generative Adversarial Networks) was used in \citep{begus_modeing_2020}. These proposals modeled perception and categorization, in relation to language acquisation. Most interestingly, representation learning has been applied for discovering phonemic tone contours in tonal languages\citep{li_representation_2020}, of which a great portion are Sinitic Languages. However, these proposals again rarely address issues from historical phonology. 

Finally, it should be noted that the concept of transforming porous data in a regular, matrix-like form to a loose, graph-like form for flexibility in processing, while essential to the designs of this paper, is not novel in the literature. Rather, it originates with the GRAPE framework in \cite{you2020handling}. Notably, when the data in question concerns Chinese historical phonology, it coincides with Johann-Mattis List's proposals for introducing network methods into computational linguistics and Chinese historical phonology. 
Generally, this line of work should be considered most relevant to our study \cite{List2018MoreON, List2014UsingPN, List2015NetworkPO}.  \citet{List2018MoreON} approaches issues spanning character formation, Middle Chinese annotation, as well as Old Chinese reconstruction with network methods.   \citet{List2014UsingPN, List2015NetworkPO} examines dialect evolution with display graphs, with a focus on the complex word-borrowing dynamics between the dialect families. He calls for colleagues to lend more attention to data-driven, quantitative methods. Our proposal answers List's call by bringing together knowledge graphs with Chinese historical phonology. Furthermore, the utilization of SOTA representation learning extends beyond the scope of the aforementioned work.

%In conclusion, strengthening the status of computational or neural methods in historical phonology is an emergent field worthy of attention.

\section{Method}

The graph-based method for representing dialect data has the benefit of making the model more flexible, robust, and efficient at using porous, incomplete data. This is particularly important since investigations into dialects are often uncoordinated, resulting in a large amount of partial character entries, where only some columns have pronunciations while others are missing. It could be argued that we can use missing data imputation to alleviate the issue, and continue processing the dialect data in a matrix form, perhaps with feed-forward neural networks or denoising autoencoders\cite{Vincent2008ExtractingAC}. However, traditional missing-data imputation techniques may create fictitious syllables that violate the phonotactics of that dialect when imputing initials or finals according to the mode of a type. Conditioning the initials or finals on each other will cause higher-order dependencies that are hard to solve. Therefore, by keeping the spaces untouched and using paired comparisons, the graph formalism circumnavigates the problem. This formulation may also allow for auxiliary input features, such as basic phonological knowledge about the nature of phonemic contrast, to be injected into the model. On this graph, we learn the embeddings with the BoxE algorithm, to be discussed below.

\subsection{Construction of a Multi-Dialectal Knowledge Graph}

\begin{figure}[ht]
\centering
\includegraphics[width=0.5\textwidth]{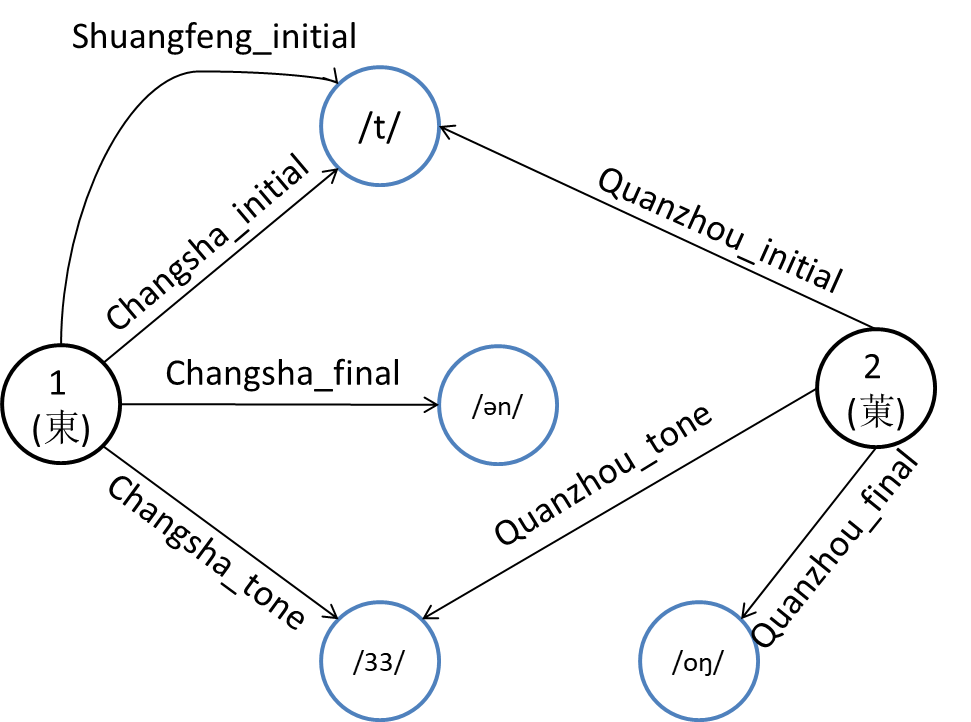}
\caption{Partial Illustration of the Phonology Knowledge Graph. The numerals represent the indices representing the Chinese characters and the glyphs for what they represent. /33/ is a tone in Chao's notation. The other nodes are segments represented in the International Phonetic Alphabet. The text labels for the edges demonstrate the how edges are categorized according to both dialect and phone type. Note that it is bi-partite by nature, as edges can only occur between “phonemic” nodes and “character” nodes, colored blue and black in the figure.(This is not provided explicitly) }
\label{GraphConstruction}
\end{figure}

%This kind of graph can be exploited when the data from every dialect are partial, and even when data for some character in an individual dialect is partial. 

We expressed the data with a knowledge graph and trained the representations through an auxiliary task of completing the multi-dialectal knowledge graph. With a graph-based technique, the representations can be more robust to noisy and porous data. Additionally, the method will be more flexible, allowing for auxiliary input features to be injected. 

We construct a graph by leveraging the characters, as well as individual initials, finals and tones from various dialects as nodes. (See Figure \ref{GraphConstruction}) .For instance, the fact of character C having an initial I in dialect D is modeled with an edge from C to I. The edge has type specific to the dialect D and the category of the component, which is an initial. This edge type can be denoted as “D-initial”. Demonstrated in Fig. \ref{GraphConstruction}, C could be character No. 1, when I is /t/ and the edge is "Changsha\_initial".

After constructing the graph, character-level and component-level representations are trained simultaneously. The knowledge graph algorithm attempts to model the nodes features as well as a prediction function so that, when given a character node and a type of link, the corresponding pronunciation node can be predicted with maximum likelihood. In this process, the model implicitly generates hypotheses about character pronunciations missing or unseen in training, as well as historical relationships between the syllables.

If there are $M$ characters with readings from $N$ dialects involved in an experiment, the upper bound for the number of edge types will be $3N$. Assuming that $F_1+F_2+F_3$ unique initials, finals and tones could be found within the aggregated phonological systems of the $N$ dialects, the upper bound for number of nodes is $M+F_1+F_2+F_3$. The graph size scales sub-linearly with the number of dialects, since as more dialects are considered, their phonemic inventories will start to overlap and exhaust.

Following convention in knowledge base research, the graph is presented in Triples of Head-Relation-Tail format. 

\subsection{The Box Embedding Model}

 In pilot tests, We considered various algorithms from the field of graph representation learning and knowledge base completion for application. In the process, it is revealed that few algorithms are inherently suitable, as there are many subtle requirements in this context:
\begin{enumerate}
\item Models designed for knowledge graphs are more suited to this application than general graph learning algorithms, since the graph to be processed is heterogeneous, besides carrying edge type as information. 
\item The model must have strong capacity for modeling multiple unique relations between the same two nodes. It is very common for one character to have the same initial across different dialects. This rules out many translation-based models, that, when given different relations, always predict different tail nodes. Prominent examples of such models include TransE \cite{bordes_translating_2013} and RotatE \cite{sun_rotate_2019}.
\item If the model uses inverse triples as an augmentation technique, then the model should also be expressive in many-to-one and one-to-many relations, because one initial or final will be mapped to numerous characters. 
\item Of the applicable algorithms, interpretability should be prioritized, since we hope to extract interpretable phonological knowledge from the obtained representations. This casts doubt on a another large family of knowledge graph models, namely the bi-linear models, epitomized by RESCAL\cite{nickel_three-way_nodate} and DistMult\cite{yang_embedding_2015}.
\end{enumerate}

After consideration, we chose BoxE for its expressiveness and tolerance to many-to-one relationships, due to its Box embedding designs. Empirically, we also demonstrate that the BoxE is relatively optimal for the phonological task through comparison with RotatE \citep{sun_rotate_2019} and ComplEx \citep{trouillon_complex_2016} in Table \ref{tab:algos}.

Here is a brief description of the BoxE algorithm. It is a translational model that embeds each node with two vectors: $e_i$, which represents the position vector, and $b_i \in \textbf{R}^d$, which represents the translational bump. These vectors are obtained after incorporating triples into the model. Additionally, each edge type is defined with two hyper-rectangles $r^{(1)}$ and $r^{(2)} \in \textbf{R}^d$. 
To satisfy the relation $R$ between entity $E_1$ and $E_2$, there is $e_1+b_2 \in r^{(1)}$ and $e_2+b_1 \in r^{(2)}$. Intuitively, this means that $E_1$ and $E_2$ "bump" each other in hyperspace $\textbf{R}^d$ by some distance. If the new vectors fall within the bounds of the associated boxes, then the proposition is considered probable. To facilitate gradient descent, the boxes have relaxed borders. It is worth noting that BoxE is also capable of hyper-graph learning as it accepts higher arity relations as input, though we did not exploit this feature for this study.

Our training objective was to maximize the score or probability of given relations. To elaborate, this means maximizing the chance of predicting masked initials/finals/tones of some character in some dialect with the unmasked components associated with that character, from both within and without the dialect. This is analog to the comparative method in Historical Phonology, as the model implicitly reconstructs a latent "proto-language", from which the descendent languages can be deduced (or, "decoded") with maximum likelihood.

\section{Data and Experimental Setup}
We use pronunciation data from four varieties of Xiang Chinese Changsha \zh{長沙}, Shuangfeng \zh{雙峰}, Guanyang Wenshi \zh{灌陽文市}, and Quanzhou Xiancheng \zh{全州縣城}., spoken primarily in Hunan Province, provided by CCR\cite{2011ccr}, and retrived with Comparative analysis toolset for Chinese dialects\cite{huang2021sinetym}. 
We also obtain labels of Middle Chinese readings from the same source. In this work, Middle Chinese refers to the phonological system recorded in the dictionary Qieyun, from the year 601 AD. It was supplemented in the Song Dynasty into the dictionary Guangyun, from which this study draws data. Middle Chinese is literary and may not reflect the colloquial speech of China in any time or place. However, most phonological systems of modern Sinitic languages (with the notable exception of the Min Languages) can be derived from the Qieyun system. Thus we treat it as a useful protolanguge model for most Sinitic Languages.

We operate on symbolic abstractions instead of raw acoustic data, as all the data have been transcribed into IPA in the database. One row of data corresponds to readings of one Chinese character. Internally, each character is mapped to a unique identifier, which is the character’s serial number in Guangyun. For every variety of Chinese, there are four columns, corresponding to initial value, final value, tonal value and tonal type of a given character’s pronuciation. 
The tone type argument is actually redundant, and it is assigned manually by investigators. In each dialect, there is a one-to-one correspondence between one tone value with one tone type. Between two dialects, tones arising from the same Middle Chinese tone are given same names. Hence, the tone type feature introduces prior expert knowledge about the historical origin of tones. However, we expect the model to derive the historical tones without any diachronic expert knowledge. Hence, we discard the tone type feature, and use only the three values for this study.

%The tone type feature is discarded for this study, as it introduces excessive cross-dialectal alignment.

\subsection{Processing of Duplicate Data}
Characters in Sinitic can be polyphonic, that is, sometimes a character will be mapped to multiple readings in one dialect. This results in duplicate data in the dataset. 
For convenience, we drop the extra pronunciations and keep only the first line for every entry. Though, there can be ambiguity surrounding the correspondence of readings for polyphonic characters. For instance, the first reading entry for a polyphonic character in dialect A might be cognate with the second reading entry for the character in dialect B. However, our naïve approach will match all the first entries to each other. Additionally, two dialects may inherit only partial readings of a polyphonic character in the proto-language. Hence, this procedure potentially introduces erroneous alignment into the model.

\subsection{Split of Training, Testing and Validating Datasets}
The model was not trained with all the data, so as to examine the robustness of the model. Instead, some triples are diverted to form testing and validating datasets. Unfortunately, assignment in this context is slightly more complicated than simple stochastic choice. There is the scenario where all initial (final/tonal) information about one character is diverted from training. In this case, the model will not be able to correctly embed this character. To circumvent this issue, we mandate that at least one feature from any of the three compositional types is retained in the training set for any character. 
In the four Xiangyu in this case, the result is empirically a split of 80.50\%:12.52\%:6.98\%.
\subsection{Data Statistics}
The initials, finals and tones count for the four dialects are listed in Table \ref{tab:data}. A total of 2805 characters is included, but not every character has the corresponding phonological data documented in every dialect. In the training set, there are 22300 entries.

\subsection{Model Setup}
For the parametric size of the model, see Table \ref{tab:model_params}. We employ the BoxE algorithm implemented in the Python library PyKeen \cite{pykeen2021software, pykeen2021benchmarking}. We did not fine-tune the model or any model parameters, so as to demonstrate the capability of the model in even in a highly suboptimal setting.

\begin{table}[h]
\centering
\begin{tabular}{l c c c}
 & \textbf{Initials} & \textbf{Finals} & \textbf{Tones} \\
 \hline
Changsha & 21 & 38 & 11 \\
Shuangfeng & 28 & 35 & 11 \\
Guanyang &  28 & 42 & 5 \\
Quanzhou & 26 & 43 & 4 \\
\end{tabular}
\caption{Data Statistics}
\label{tab:data}
\end{table}

\begin{table}[h]
  \centering
  \begin{tabular}{lc}
    \toprule
    \textbf{Parameter} & \textbf{Value} \\
    \midrule
    Vector and hyperbox dimension & 64 \\
    Number of nodes & 2946 \\
    Number of edge types & 12 \\
    Cumulative parameter size & 378624 \\
    Optimization algorithm & Adam \\
    Number of epochs & 2000 \\
    \bottomrule
  \end{tabular}
  \caption{Model Parameters}
  \label{tab:model_params}
\end{table}

\section{Experimental Evaluation}
\begin{figure}[ht]
\centering
\includegraphics[width=0.5\textwidth]{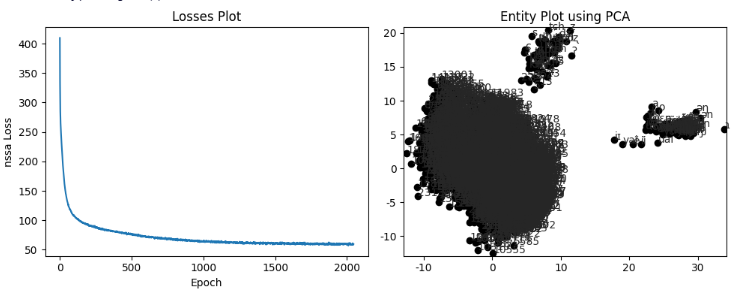}
\caption{Preliminary Visualization of Training Dynamics and Trained Embeddings. }
\label{fig:preliminary}
\end{figure}

\begin{figure}[ht] \centering \begin{minipage}[b]{0.45\linewidth} \centering \includegraphics[width=\textwidth]{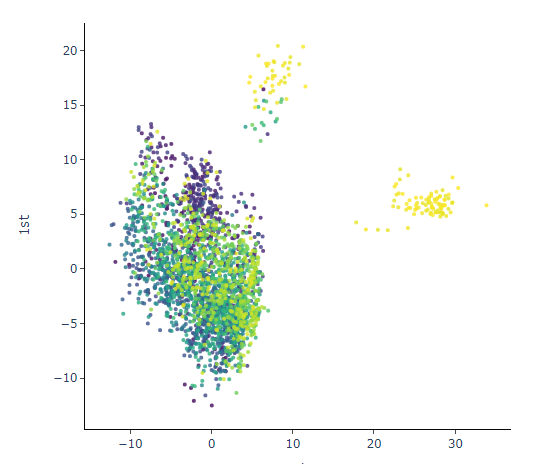} \end{minipage} \hspace{0.5cm} \begin{minipage}[b]{0.45\linewidth} \centering \includegraphics[width=\textwidth]{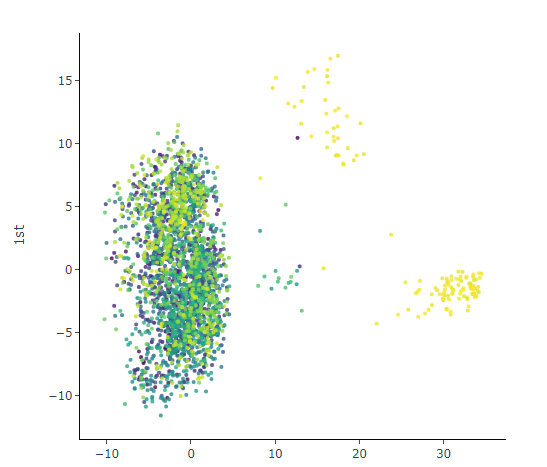} \end{minipage} \caption{UMAP(\citealp{mcinnes2018umap-software}, \citealp{2018arXivUMAP},Uniform Manifold Approximation and Projection) decomposed visualizations of the translational bumps (a) and position embeddings (b). The coloring reflects a point's index in the Guangyun, which is sorted according to rhyme.} \label{fig:decomp} \end{figure}

\subsection{Canonical Evaluation of Model}
The convergence of the model, and a preview of the spatial distribution of embeddings can be seen in Figure \ref{fig:preliminary}. The model quickly converges. The entity plot decomposed with PCA reveals a mass of character readings “ejecting” two groups of entities, respectively the combination of all initials and tones, and all finals, which is in accordance with the bi-partite and heterogeneous nature of this graph.

Canonically, BoxE is evaluated with the hit@n metric and MRR (mean reciprocal rank) for link prediction. On the validation set, our model achieved hit@1:51.25\%, hit@5: 87.19\%, hit@10: 93.76\% on the “tail” batches. The head batches are not relevant because they involve “predicting characters from initials/finals”, of which there is many to one. In Table \ref{tab:algos}, we demonstrate empirically the superiority of the BoxE algorithm over other common knowledge graph algorithms on this phonological task. 
A clearer visualization of the embedded points can be seen in Figure \ref{fig:decomp}. Guangyun ensures that rhyming characters (having the same final) have similar coloring on the map. The coloring is only a reflection of the point's serial in the dataset and does not have any quantitative interpretation. Presumably, the translational bump for characters will contain more relevant information to historical phonology, as they designate which component types to "bump into the box." Without mention, all experiments are carried out on the bump embeddings and not positions. However, empirically we find that the two kinds of embeddings are interchangeable.

\subsection{Examining Contrastive Information}

In this section, unsupervised clustering is used to evaluate contrastive information in the embeddings.
Based on the hypothesis that the phonological structures of the dialects are co-embedded in the latent structure of embeddings, we determined if the high-dimensional embeddings retain information associated with the theoretic categories of the input dialects, a similar task to \citealt{tilsen_localizing_2021}. After applying a clustering algorithm to the embedded characters, the information yield \footnote{Entropy subtracted by conditional entropy, or an empirical estimate of mutual information.} of the found categories against input categories of initials, finals and tones is computed.  A higher information yield indicates that the clusters found by unsupervised clustering were more interpretable with respect to the input phonemic categories. 
\footnote{HDBSCAN sometimes refuses to classify points it is not sure of. These points are combined into one category for the aforementioned purpose.}
\footnote{Before using HDBSCAN, UMAP was first used to reduce the 64 embedding dimensions to 8 dimensions, with the neighbour parameter set to 50. This is an advised practice from the HDBSCAN documentation.}

\begin{figure*}[ht]
\centering
\includegraphics[width=0.9\textwidth]{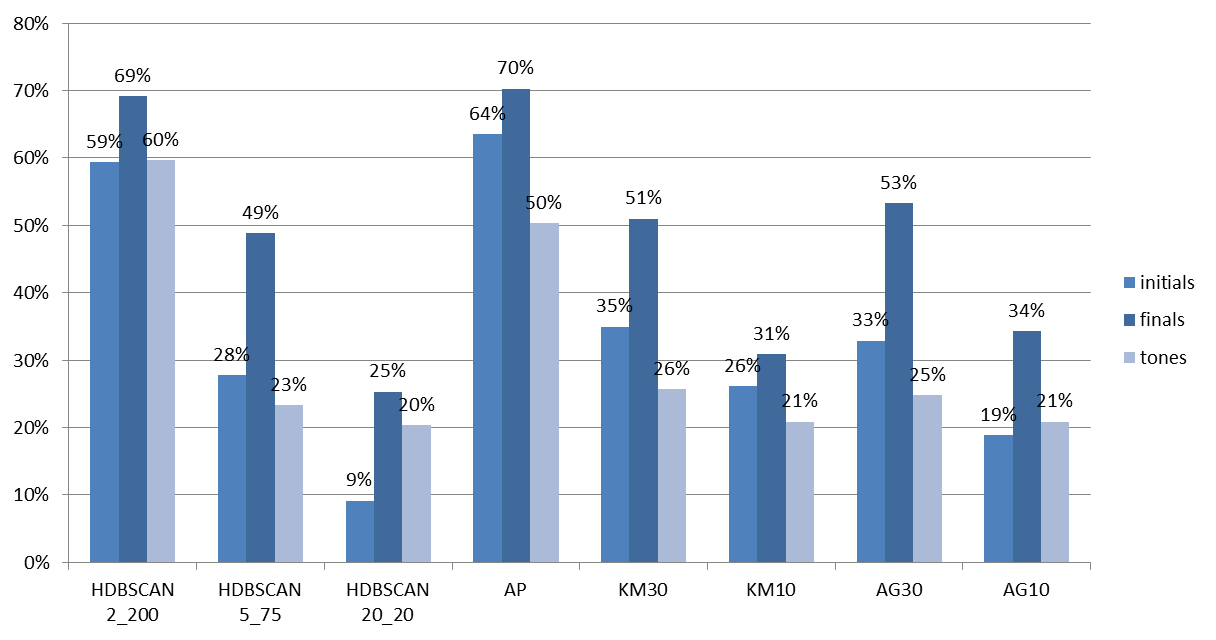}
\caption{Information yield in percentage averaged across four dialects.  For HDBSCAN, the min samples and min cluster size parameters were set to 2 and 200, 5 and 75, 20 and 20 respectively. The other three methods were employed on the original embeddings. For K-means and agglomerative clustering, the number of clusters was specified to be 30 and 10.}
\label{fig:information}
\end{figure*}

The clustering algorithms used for dissecting the cloud of embedded characters include HDBSCAN (\citealp{mcinnes2017accelerated},A density based method), Affinity Propagation, K-means and Agglomerated Clustering.\footnote{The numerous methods were tried sequentially as we do not know which algorithm best recovers the latent structure of representations in accordance with theoretic categories.} The results can be seen in Figure \ref{fig:information}.

Affinity propagation and HDBSCAN achieved best effects on finding interpretable clusters from the datasets. Though, we find that HDBSCAN is very sensitive to the two parameters: its effect degrades when we allow for smaller clusters but demands greater confidence on the classification. Notably, HDBSCAN achieved an effect similar to affinity propogation with just 29 clusters, while the latter used 130. 

The large information yields reflect that the unsupervised algorithms do tend to dissect the character set along latent lines corresponding to phonological opposition in the input dialects, as shown in a partial observation in Table \ref{tab:hdbscan_clusters}.
It appears that the distribution of finals in dialects had more influence on the latent structure than initials or tones. Simply put, the characters within each unsupervised cluster are more likely to rhyme than alliterate, though both cases occur in observation of the HDBSCAN Clusters.

%\begin{figure}[ht]
%\centering
%\includegraphics[width=0.5\textwidth]{HDBSCAN2020.png}
%\caption{UMAP visualization of Character Embeddings, coloured %with categories from HDBSCAN.}
%\label{fig:umap}
%\end{figure}

\begin{table*}[htbp]

\centering

\begin{tabular}{cccccc}
\hline
\textbf{ID} & \textbf{Changsha} & \textbf{Shuangfeng} & \textbf{Guanyang} & \textbf{Quanzhou} \\
\hline
0 & Initial:/m/ & Initial:/m/ & Initial:/m/ & Initial:/m/ \\
1 & Initial:/p\textsuperscript{h}/ & Initial:/p\textsuperscript{h}/ & Initial:/p\textsuperscript{h}/ & Initial:/p\textsuperscript{h}/ \\
2 & Final:/{\~i}n/ & Final:/{\~i}/ & Final:/ i\textipa{\~E}/ & Final:/ ie\textipa{ŋ}/ \\
7 & Final:/(u)ei/ & Final:/ui/ & Final:/ u\textipa{E}i/ & Final:/uei/ \\
\hline
\end{tabular}
\caption{Analysis of Selected HDBSCAN Clusters. In these clusters, characters are predominantly, but not exclusively associated with the listed features. }
\label{tab:hdbscan_clusters}

\end{table*}
 
There are limitations to this experiment though, which will be discussed below.

\subsection{Inference of Proto-language Features }
In this section, we investigate the quality of our embeddings with respect to proto-language reconstruction tasks, as an important potential application of this method lies with such work. Hence, we trained classifiers in attempt to infer labels from Middle Chinese, which likely predates proto-Xiang, therefore an accessible surrogate for that proto-language.

The features to infer are Grades (\zh{等地}), Voice(\zh{清濁}), Tones(\zh{聲調}), She (\zh{攝}, a coarse division of finals), Initials (\zh{字母}), and Mu(\zh{韻目},a fine division of finals).

Grades are believed to be associated with medials, a component in the front of the final (amalgamated with final in Xiangyu data). Voice is a division based on properties of the initial, in which voiced consonants, voiceless unaspirated consonants, voiceless aspirated consonants and nasal consonants are distinguished. For tones, in Middle Chinese, there were four: level, rising, departing, and entering.
Of these categorical labels, there are respectively 4, 4, 4, 16, 36 and 206 unique classes. \footnote{Canonically so, but there are a few erroneous entries in the data we used, resulting in sometimes one or two extra categories containing a few characters. They were kept.}

For this experiment, a train-test split of 0.67-0.33 was instated. Since phonological evolution is quite regular and systematic, we should expect decent results without a great proportion of data used for training. Accuracies below are for the test set.
These values are consistently higher than a naïve baseline of guessing the mode of each distribution, proving that proto-language related features were preserved in the retrived embeddings. (See Table \ref{tab:proto}.)

\begin{table}[t]
    \centering
    \begin{tabular}{|l|c|c|c|}
        \hline
        \textbf{Alg. (Metric \%)} & \textbf{Hit@1} & \textbf{Hit@5} & \textbf{Hit@10} \\ \hline
        \textbf{BoxE} & \textbf{51.25} & \textbf{87.19} & \textbf{93.76}  \\
        \textbf{RotatE} & 33.11 & 57.47 & 66.18 \\
        \textbf{ComplEx} & 9.40 & 24.65 & 35.37 \\
       \hline
    \end{tabular}
    \caption{An empirical demonstration of the superiority of the BoxE algorithm for the phonological investigation task among common missing link prediction methods. The models were set to the same embedding dimension. None of the models were fine-tuned or ran for more than a single time, hence all readings should be seen as sub-optimal.}
    \label{tab:algos}
\end{table}

\begin{table*}[t]
    \centering
    \begin{tabular}{|l|c|c|c|c|c|c|}
        \hline
        \textbf{Algorithm(Acc \%)} & \textbf{Grades} & \textbf{Voice} & \textbf{Tones} & \textbf{She} & \textbf{Initials} & \textbf{Mu} \\ \hline
        \textbf{Ridge Classification} & 65.3 & 76.4 & \textbf{84.1} & 54.6 & 49.4 & 18.6 \\
        \textbf{MLP} & \textbf{70.5} & \textbf{81.1} & 83.0 & \textbf{61.4} & \textbf{53.2} & \textbf{26.9} \\
        \textbf{Naïve Baseline} & 48.4 & 35.4 & 35.6 & 15.3 & 8.1 & 1.8 \\
       \hline
    \end{tabular}
    \caption{Comparison of Ridge and MLP probes for proto-language Feature Inference. The baseline is the accuracy obtained by uniformly guessing the most frequent class for each character.}
    \label{tab:proto}
\end{table*}

%hit@1:51.25\%, hit@5: 87.19\%, hit@10: 93.76\%

The MLP generally outperforms Ridge Classification on inference for these characters, with the sole exception of tones, where RC outperforms MLP by 1.1\%. The best results are attained for tones and voice, showing these features to be phonologically well preserved from Middle Chinese to Xiang languages. 

\begin{figure}[ht]
\centering
\includegraphics[width=0.5\textwidth]{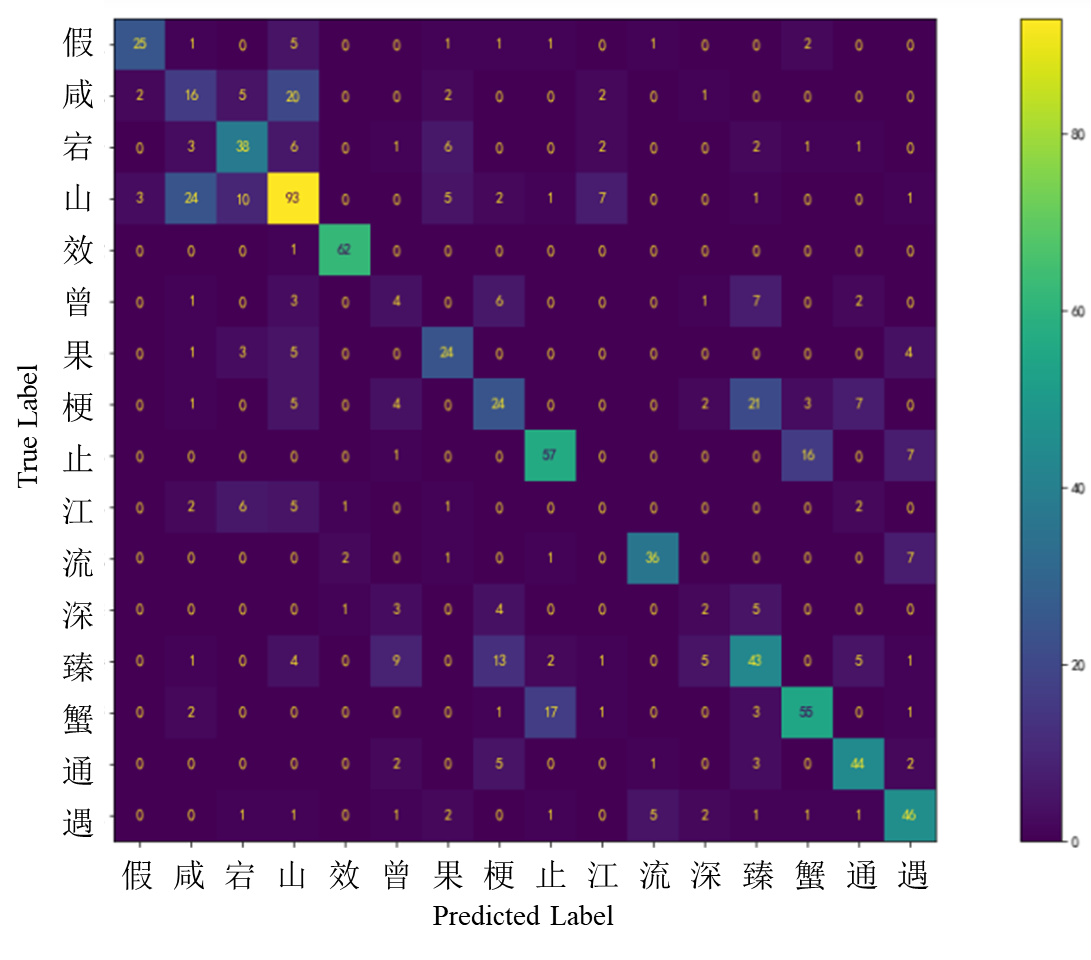}
\caption{Confusion matrix for She.}
\label{fig:confusion}
\end{figure}

Interesting observations can be drawn from the confusion matrices generated with such classification. Presumably, these matrices can offer insight into what categories were blended, which oppositions were lost during the development of some language family. One such example is demonstrated in Figure \ref{fig:confusion}. It could be seen that there is large confusion between the Xian \zh{咸}, Dang \zh{宕} and Shan \zh{山} Shes, and also between Xie \zh{蟹} and Zhi \zh{止} Shes. \footnote{In Baxter's transcription, \zh{咸} = \textit{-eam}, \zh{宕} = \textit{-ang}, \zh{山} = \textit{-ean}; \zh{蟹} = \textit{-ea}, \zh{止} = \textit{-i} \citep{Baxter2014OldCA}. There are only hypothetical IPA values available for these archaic categories.} This could indicate that in Proto-Xiang, there is confusion between these categories relative to Middle Chinese.

\section{Discussions}

Our current setting only operates on pre-abstracted symbols and lacks incorporation of acoustic or articulatory evidence. Incorporating multi-modal data into a knowledge graph framework could enhance the quality of embeddings and enable more accurate representations of phonological features. Alsp, the proposed method uses shared embeddings for symbolic components across different dialects, which cannot fully capture dialect-specific variations. Investigating contextualized or dialect-specific component embeddings could improve the model's ability to capture finer-grained phonological distinctions. Finally, phonetically similar components are currently treated as independent items, which is too absolute an assumption. However, it is also possible for phonetic cues to override the correct phonological alignment in the model. In many cases, phonetic similarity does not imply diachronic homology. Two phonetically equivalent syllables from two different dialects may have different origins. Conversely, two phonetically distinct syllables from two different dialects may be cognate.
The subtle balance between "phonetic" and "phonological" proximity requires further discussion.

Several lines of research may benefit from robust multi-dialectal representations. In dialectology, there is need for estimating divergence between phonological systems. That includes the divergences between its constituents, such as individual characters, phonemes and syllables. With multi-dialectal representations, this divergence can be estimated quantitatively. In historical phonology, the reconstruction of a proto-language demands deep scrutiny of dialect systems whose efficiency can be improved with manipulating the representations. Also, they can be used for completion of the phonological knowledge base. Often knowledge bases for Sinitic phonology are fragmented, due to imperfect surveys and heterogeneity of sources, etc. The representations can be used to infer missing pronunciations in different dialects to improve the quality of observations.

The graph-based method proposed in this paper benefits from phonological characteristics specific to Sinitic languages, but is also limited by these characteristics. Specifically, the process of constructing a phonological graph from words, as proposed in this study, is less natural in languages where words typically have many syllables, and vary in the number of syllables contained. In these languages, the temporal interaction of syllables within a word is a new phenomena that the graph-based method needs to adapt to. Additionally, in these languages, it will be less straightforward to tokenize the words into expressive sub-words to use as nodes in the graph. Presumably, in non-Sinitic languages, the proposed method will be most performant in other languages of the Southeast Asian Sprachbund, such as those in the Hmong-Mien or Austroasiatic families. These languages share phonological features with Sinitic languages that enable our method. On the other hand, this method will likely meet more complications outside of the local sprachbund.

\section{Conclusion}
This paper demonstrated the potential of graph-based representation learning in Chinese Historical Phonology. The representations are potent in many ways, i.e. facilitating the reconstruction of minor proto-languages.

 In the future, more sophisticated techniques such as deep learning models could be explored to further improve the quality of the obtained representations. Furthermore, the proposed method can be integrated with other linguistic resources, such as recordings, articulatory time series, or orthographic corpora, to enrich the knowledge base and improve the accuracy of reconstructions. With the development of modern, massive linguistic datasets such as Nk2028\cite{zhou2020nk2028}, CogNet\cite{batsuren_large_2022} or MorphyNet\cite{batsuren_morphynet_2021} as well as improvements in large pre-trained models, we can expect foundational models that possess emergent and meta-generalizing capabilities to arise in historical phonology or morphology. This avenue of research holds great promise for advancing our understanding of the phonology and evolution of Sinitic languages, and potentially other language families as well.

\section*{Limitations}
This study stems from a novel idea for Chinese Historical Phonology Studies. As few direct predecessors could offer hindsight, there are quite a few limitations to this study that may be addressed with further work. 
\begin{enumerate} \item While the initial-final-tone decomposition is convenient in this context, it also limits the transferrability of the proposed tool to languages outside of the Sinosphere. This calls for further exploration of more generalizeable approaches to phonological representation learning.

\item Polyphonic characters were not fully utilized in the study, and their alignment per-reading and tokenization into separate identifiers should be considered in future work.

\item Finally, making full use of the dataset is crucial, and the stochastic train-test split used in this study may leave out important hints. Alternative sampling strategies, such as cross-validation or bootstrapping, could enhance the robustness of the results. \end{enumerate}

\section*{Acknowledgements}
 We are grateful for the valuable advice and feedback we received from various peers during the course of this work. Without their contributions, this research would not have been possible.

\bibliography{anthology, custom}
\bibliographystyle{acl_natbib}

%\appendix

\end{document}